\documentclass[10pt,twocolumn,letterpaper]{article}
\usepackage{subcaption}

\usepackage{iccv}

\usepackage{times}
\usepackage{epsfig}
\usepackage{graphicx}
\usepackage{amsmath}
\usepackage{amssymb}
\usepackage[ruled,vlined,linesnumbered]{algorithm2e}
\usepackage{url}
\usepackage{mathrsfs}
\usepackage{multirow}
\usepackage{makecell}
\usepackage{colortbl}
\usepackage{booktabs}
\usepackage{array}
\usepackage{newfloat}
\usepackage{listings}
\usepackage[dvipsnames]{xcolor}
\definecolor{lightgray}{rgb}{0.9, 0.9, 0.9}
\definecolor{ballblue}{rgb}{0.13, 0.67, 0.8}
\newcommand{\STAB}[1]{\begin{tabular}{@{}c@{}}#1\end{tabular}}

\usepackage[pagebackref=true,breaklinks=true,letterpaper=true,colorlinks,bookmarks=false,citecolor=ballblue]{hyperref}

\iccvfinalcopy 

\def\iccvPaperID{1535} 

\ificcvfinal\fi

\begin{document}

\title{Revisiting the Parameter Efficiency of Adapters from the Perspective of Precision Redundancy}
\author{Shibo Jie\qquad Haoqing Wang\qquad Zhi-Hong Deng\thanks{Corresponding Author.}\\
School of Intelligence Science and Technology, Peking University\\
{\tt\small \{parsley,wanghaoqing,zhdeng\}@pku.edu.cn}
}

\maketitle
\ificcvfinal\thispagestyle{empty}\fi

\begin{abstract}

Current state-of-the-art results in computer vision depend in part on fine-tuning large pre-trained vision models. However, with the exponential growth of model sizes, the conventional full fine-tuning, which needs to store a individual network copy for each tasks, leads to increasingly huge storage and transmission overhead. Adapter-based Parameter-Efficient Tuning (PET) methods address this challenge by tuning lightweight adapters inserted into the frozen pre-trained models. In this paper, we investigate how to make adapters even more efficient, reaching a new minimum size required to store a task-specific fine-tuned network. Inspired by the observation that the parameters of adapters converge at flat local minima, we find that adapters are resistant to noise in parameter space, which means they are also resistant to low numerical precision. To train low-precision adapters, we propose a computational-efficient quantization method which minimizes the quantization error. Through extensive experiments, we find that low-precision adapters exhibit minimal performance degradation, and even 1-bit precision is sufficient for adapters. The experimental results demonstrate that 1-bit adapters outperform all other PET methods on both the VTAB-1K benchmark and few-shot FGVC tasks, while requiring the smallest storage size. Our findings show, for the first time, the significant potential of quantization techniques in PET, providing a general solution to enhance the parameter efficiency of adapter-based PET methods. Code: \url{https://github.com/JieShibo/PETL-ViT}

\end{abstract}

\section{Introduction}
Large pre-trained vision models have demonstrated exceptional performance on various visual tasks via fine-tuning on task-specific data. In the traditional fine-tuning paradigm, the entire model is updated for each downstream task, resulting in the need to store a fine-tuned model separately for each task. However, with the remarkable scalability of modern vision models, the size of pre-trained vision models is increasing exponentially to achieve superior performance. As a result, the storage cost of the full fine-tuning paradigm becomes prohibitive in multi-task scenarios.

\begin{figure}[t]
     \centering
         \includegraphics[width=0.4\textwidth]{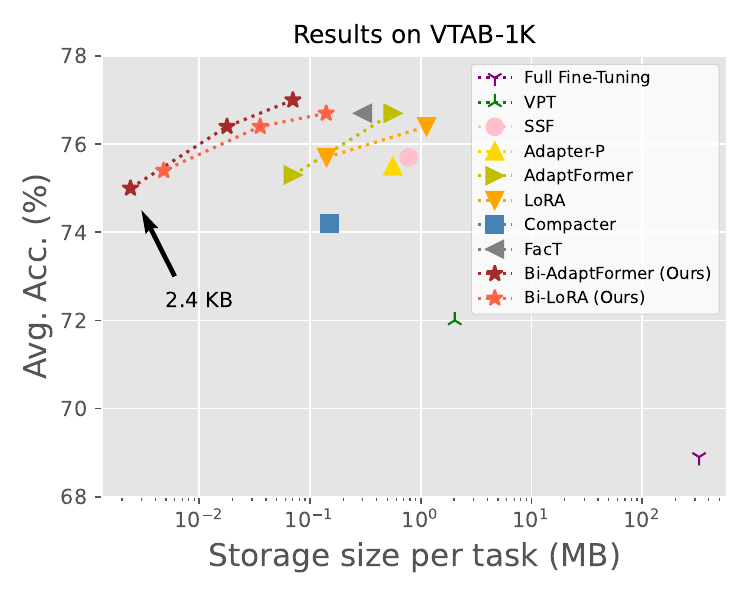}
         \caption{\textbf{Average accuracy \emph{vs.} size of trainable parameters in backbones (log scale) on VTAB-1K benchmark.} Our low-precision adapter-based methods outperform other baselines.}
         \label{fig:intro}
         \vspace{0pt}
\end{figure}

\emph{Parameter-Efficient Tuning} (PET) has recently emerged as a promising approach for fine-tuning a limited number of parameters while attaining performance comparable to full fine-tuning on downstream tasks. Adapter-based methods~\cite{adapter-cv,adapter,adapterp,adaptformer,compactor,convpass,lora,fact} are among the techniques proposed for PET and have gained considerable attention due to their effectiveness. Adapters are typically small subnetworks with bottleneck architecture comprising two fully-connected (FC) layers inserted into pre-trained models. Adapter-based methods freeze pre-trained weights and update only the adapters, whose parameter efficiency is achieved through their small hidden dimension.

Although the bottleneck adapters have been already lightweight (\eg, 0.5 MB/task for ViT-B~\cite{vit}), the storage costs remain considerable when dealing with a huge number of tasks (\eg, platform that provides customized models for millions of users). To address this issue, recent studies have shown that the parameter efficiency of adapters can be further improved. For example, \cite{compactor,fact,pevit} explore the low-rank structure in adapters, reparameterizing the weight of adapters into smaller subspace with Kronecker, Tensor-Train, or Tucker factorization. Additionally, \cite{sparseadapter} leverages network pruning to train sparse adapters. We find that these methods actually have a common motivation -- reducing the redundancy (\eg, rank redundancy, density redundancy) in adapters. Also motivated by this, we pose a question, \emph{whether there is any other kind of redundancy that can be utilized to better improve the efficiency of adapters.}

In this paper, we begin by exploring the loss landscape of adapters and observe that the local minima of adapters are much flatter than that of the fully fine-tuned models. The flatness of local minima indicates that the trained adapters possess greater resilience to noise in parameter space, such that adapters with low-precision parameters should perform equally well as their high-precision counterparts. Therefore, we infer that adapters are redundant in numerical precision. Since previous work on adapters all employs full-precision (FP32) data type, the impact of precision on adapters has not been investigated yet.

To reduce the precision redundancy, we propose an approach that involves training and storing adapters in low-bit parameter space. Through empirical analysis, we observe that the parameters of each adapter weight approximately follow a Gaussian distribution. Under this assumption, we quantize the adapter parameters by minimizing the quantization loss. Inspired by previous work of neural network quantization~\cite{quansurvey}, we adopt quantization-aware training and train the low-bit adapters with straight-through estimator (STE). Our experiments, conducted on extensive datasets, reveal several key findings: \emph{1)} Unlike quantizing the entire model, quantizing only the adapters results in negligible performance degradation, even in the 1-bit setting; \emph{2)} With a fixed storage budget, 1-bit quantized (\ie, binary) adapters achieve superior  performance among all precision settings; \emph{3)} Our 1-bit adapter can outperform all previous PET methods, including low-rank factorization methods, while using the smallest  storage size.

Our contributions are summarized as follows:
\begin{itemize}
    \item From the investigation on the flat local minima of adapters, we infer the existence of precision redundancy in the parameters of adapters, which can be leveraged to improve their parameter efficiency.
    \item Based on empirical observations of the distribution of adapter parameters, we propose an efficient quantization-aware training method for learning low-bit adapters while minimizing the quantization error.
    \item Extensive experiments and comparisons verify that lowering the bit-width brings significant efficiency improvement to adapters. Our proposed method achieves new state-of-the-art results in terms of both performance and parameter efficiency.
\end{itemize}

\begin{figure*}[t]
     \centering
         \includegraphics[width=0.8\textwidth]{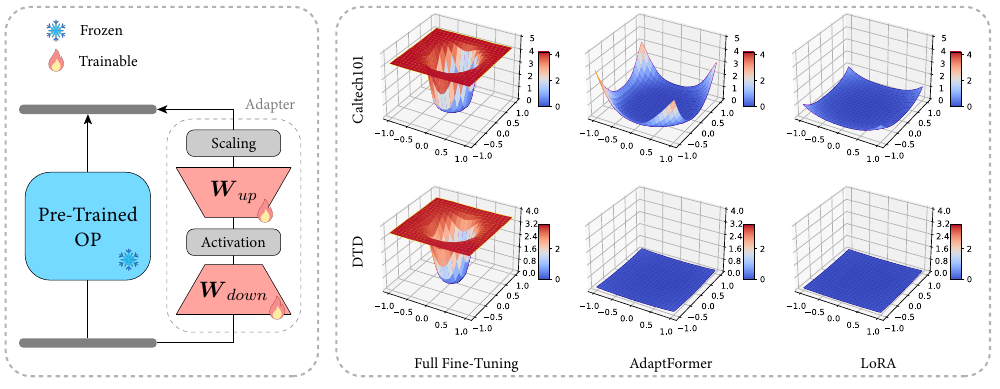}
         \caption{\textbf{\emph{Left:}} Illustration of adapters. ``Pre-Trained OP'' denotes operations in pre-trained models, such as the FFN blocks or QKV transformations in ViTs.  \textbf{\emph{Right:}} Loss landscape visualization of full fine-tuning and adapter-based tuning~\cite{adaptformer,lora} on ViT-B.}
         \label{fig:landscape}
                  \vspace{0pt}
\end{figure*}
\begin{figure}[t]
     \centering
         \includegraphics[width=0.48\textwidth]{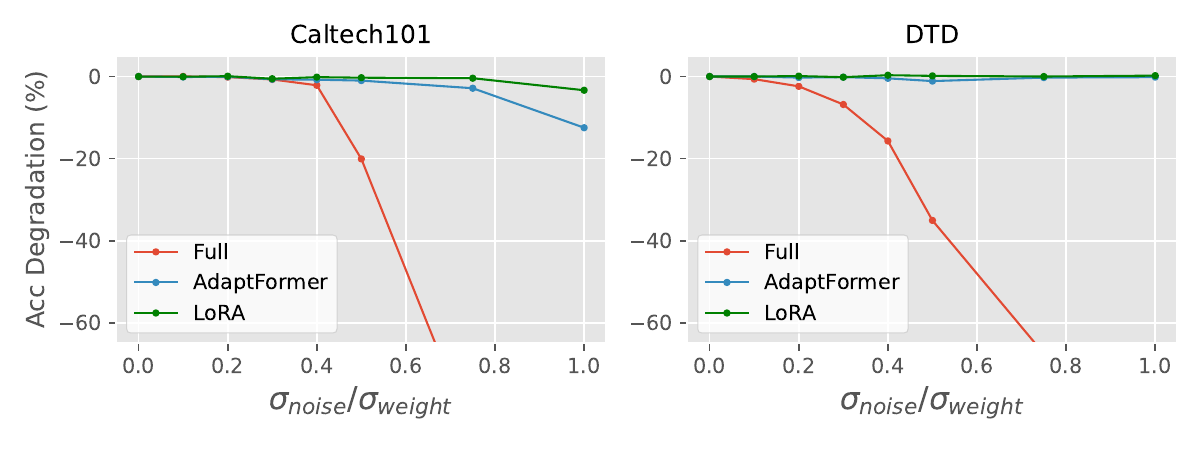}
         \caption{\textbf{Accuracy degradation under different intensity of Gaussian noise.} Adapters converge at flatter local minima and are more resistant to disturbation.}
         \label{fig:noise}
          \vspace{0pt}
\end{figure}
\section{Related Work}
\textbf{Parameter-Efficient Tuning.}
Parameter-Efficient Tuning (PET) aims to adapt pre-trained vision backbone to downstream tasks by tuning only a small number of parameters. Most work about PET focuses on tuning transformer-based networks, \eg, Vision Transformers (ViTs)~\cite{vit}. Prompt-based methods~\cite{vpt,coop,cocoop,prompt_1,prompt_2,prompt_3} concatenate trainable tokens to the sequential inputs of transformers as prompts, adapting the models by tuning the prompts. However, since the computational cost of self-attention is proportional to the square of the length of inputs, prompt-based methods are not as computation-efficient as the original network~\cite{vpt,adaptformer}. Adapter-based methods~\cite{adapter-cv,adapter,adapterp,adaptformer,compactor,convpass,lora,fact,consolidator,repadapter} insert small adapters into the pre-trained model, adjusting the intermediate representations of the network to fit the downstream data. Some of them~\cite{lora,fact} can be absorbed into the pre-trained weights during inference, which ensures the computational cost is not increased. Besides, there are also methods that tune bias parameters~\cite{bitfit}, modify the intermediate features via affine transformation~\cite{ssf}, fit the change in the network outputs by a small side-network~\cite{sidetune}, or combine multiple methods automatically~\cite{noah,glora}. Among them, adapter-based methods have attracted much attention for their competitive performance, generality to different backbones, and scalability.

\textbf{Efficient Designs of Adapters. }
As illustrated in Figure~\ref{fig:landscape} (left), adapters are commonly subnetworks composed of two FC layers with nonlinear activation in between. {\sc Adapter-P}~\cite{adapterp} places the adapters after the Feed-Forward Network (FFN) blocks, and {\sc AdaptFormer}~\cite{adaptformer} uses adapters parallel to the FFN blocks of ViT. {\sc LoRA}~\cite{lora} uses two low-rank matrices to fit the change in the query and value transformation of Multi-Head Self-Attention (MHSA). The formulation of {\sc LoRA} is equivalent to two FC layers without bias parameters and activation, and can be regarded as special adapters in parallel with the query and value weights.

Besides, some work focuses on more compact designs for adapters. {\sc Compacter}~\cite{compactor} and {\sc KAdaptation}~\cite{pevit} regard the weights of adapters as the Kronecker product of two smaller matrices, one of which is shared among adapters. {\sc FacT}~\cite{fact} tensorizes the network as a tensor, and reparameterizes its change as several factors according to Tensor-Train or Tucker format that are updated end-to-end. Similar to {\sc LoRA}, {\sc FacT} is not proposed as an adapter-based method, but it can also be viewed as reparameterized adapters with partially shared weights. Besides, {\sc SparseAdapter}~\cite{sparseadapter} prunes the dense weights of adapters before fine-tuning. These designs reduce the rank and density redundancy in adapters, but we focus on a neglected but more effective direction -- precision redundancy.

\textbf{Network Quantization. }
Network quantization~\cite{quansurvey} compresses networks by reducing the bit-width of weight and activation. Current quantization methods include Post-Training Quantization~\cite{ptq1,ptq2,ptq3,ptq4,ptq5,ptq6,ptq7}, which performs quantization on trained model without re-training; and Quantization-Aware Training~\cite{qat1,qat2,qat3,qat4}, which introduce quantization during the training process by approximating the gradient of the non-differentiable quantization operator. The former paradigm does not require access to the entire training data during quantization and has shown almost lossless performance using FP16 and INT8 data type, while the latter yields quantized models with better performance and can work in extremely low-bit settings, \eg, binary quantization~\cite{bnn1,bnn2,bnn3}.

\section{Preliminaries}
In this paper, we mainly focus on ViTs as pre-trained backbone following previous work~\cite{vpt,adaptformer,noah,convpass}. We start with a concise formalization of the commonly used adapters.

\textbf{{\sc AdaptFormer}}~\cite{adaptformer} uses bottleneck FFN composed of two FC layers with in-between ReLU activation as adapters. The weights of an adapter are $\boldsymbol{W}_{down}\in \mathbb{R}^{d\times h}$ and $\boldsymbol{W}_{up}\in \mathbb{R}^{h\times d}$, where $h << d$. Adapters are inserted into networks as shortcuts bypassing the FFN blocks, \ie, given an input $\boldsymbol{X}\in \mathbb{R}^{N\times d}$, the computation is  formulated as
\begin{equation}\boldsymbol{X}'=\underbrace{\boldsymbol{X}+\textit{FFN}(\boldsymbol{X})}_{\text{Frozen}}+\underbrace{s\cdot\textit{ReLU}(\boldsymbol{X}\boldsymbol{W}_{down})\boldsymbol{W}_{up}}_{\text{Adapter}}\end{equation}
where $s$ is a hyper-parameter, $\boldsymbol{X}$ is the input of FFN blocks.

\textbf{{\sc LoRA}}~\cite{lora} learns the low-rank approximation of change in $\boldsymbol{W}_{q}$ and $\boldsymbol{W}_{v}$. Formally, it reparameterizes $\Delta\boldsymbol{W}_{q/v}$ into $\boldsymbol{A}_{q/v}\boldsymbol{B}_{q/v}$, where $\boldsymbol{A}_{q/v}\in \mathbb{R}^{d\times h}, \boldsymbol{B}_{q/v}\in \mathbb{R}^{h\times d}$ and $h << d$. The query and value of MHSA are computed as
\begin{equation}
\boldsymbol{Q/V}=\underbrace{\boldsymbol{XW}_{q/v}}_{\text{Frozen}}+\underbrace{s\cdot\boldsymbol{XA}_{q/v}\boldsymbol{B}_{q/v}}_{\text{Adapter}}\end{equation}
in which $s$ is a scaling hyper-parameter, and $\boldsymbol{X}$ is the input of MHSA blocks. {\sc LoRA} is equivalent to using {\sc AdaptFormer}-style adapters with identity activation, whose weights are $\boldsymbol{A}_q,\boldsymbol{B}_q,\boldsymbol{A}_v,\boldsymbol{B}_v$.

\begin{figure*}[t]
     \centering
         \includegraphics[width=0.98\textwidth]{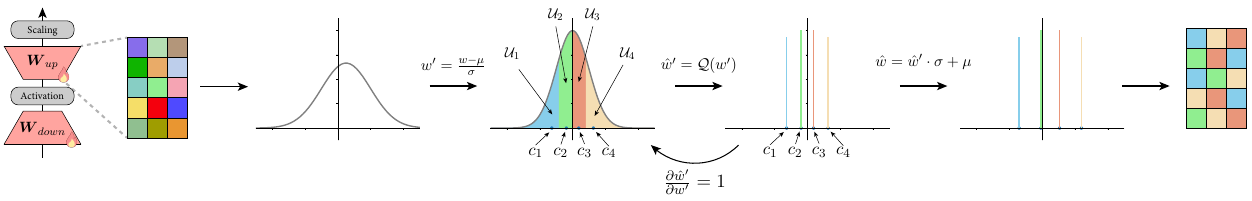}
         \caption{\textbf{Illustration of the proposed quantization method with $b=2$.}}
         \label{fig:quantization}
                  \vspace{0pt}
\end{figure*}
\begin{figure}[t]
     \centering
         \includegraphics[width=0.48\textwidth]{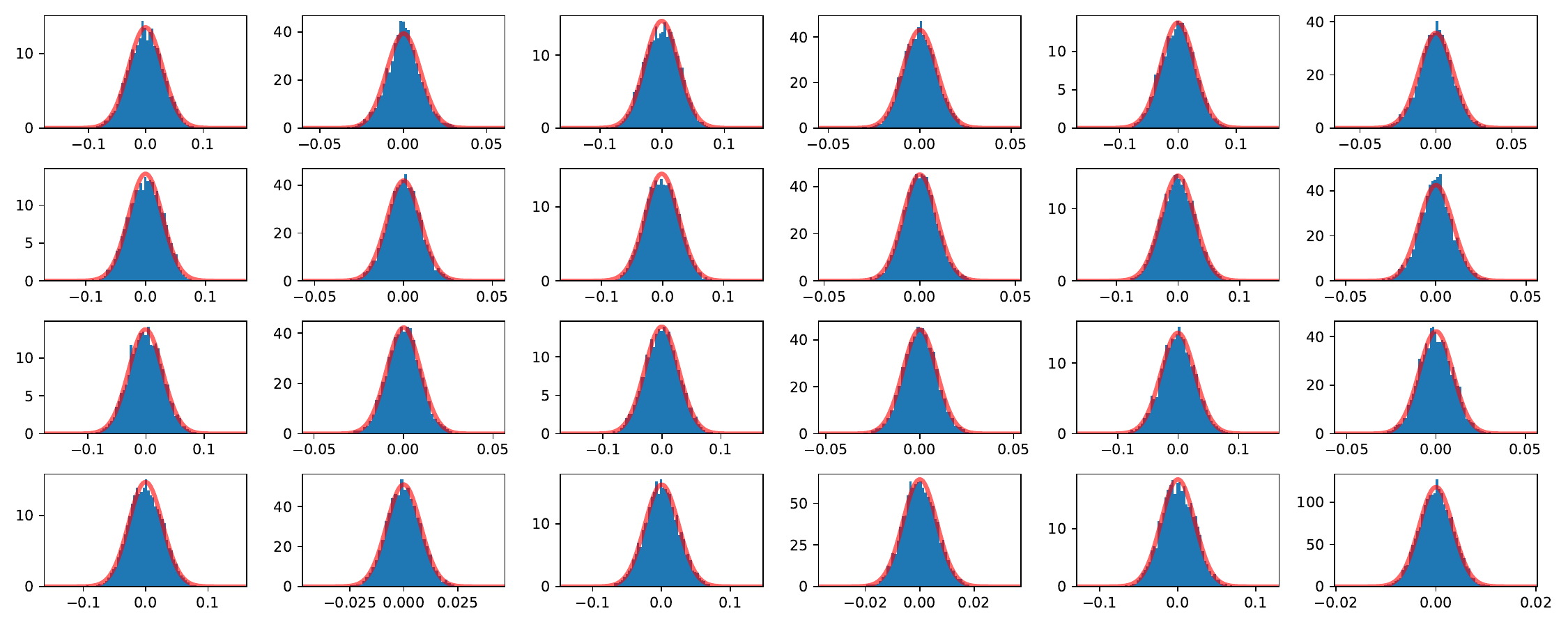}
         \caption{\textbf{Parameter frequency histogram visualization} of the 24 weight matrices in all the 12 adapters of {\sc AdaptFormer} fine-tuned on Caltech101. The parameters (blue histograms) are roughly subject to Gaussian distribution (red curves).}
         \label{fig:histogram}
          \vspace{0pt}
\end{figure}

\section{Methodology}
\subsection{Precision Redundancy in Adapters}
It has been extensively studied that the property of a neural network is highly correlated with the flatness of its loss landscape, \eg, the flatter the local minima, the better the generalization~\cite{landscape1,landscape2,landscape3,landscape4,whenvit,effectiveness}. 
Inspired by them, we here investigate the loss landscape of adapters in vision models to explore their property. Following~\cite{landscape1}, we plot the loss landscape of full fine-tuning, {\sc AdaptFormer}, and {\sc LoRA} when adapting pre-trained ViT-B~\cite{vit}. As shown in Figure~\ref{fig:landscape} (right), {\sc AdaptFormer} and {\sc LoRA} obviously converge at much flatter regions than full fine-tuning.


The flat local minima of visual adapters indicate that they generalize better, providing an explanation for their superior performance over full fine-tuning on small and medium-size datasets~\cite{vpt,convpass}. Moreover, 
if the parameters converge at flatter local minima, there are wide low-loss areas around these points. Therefore, when adding noise to the converged parameters, we can expect that the loss will not increase significantly. In other words, the model is resistant to disturbation in parameter space. 

As shown in Figure~\ref{fig:noise}, we add Gaussian noise $\mathcal{N}(0,\sigma_{noise}^2)$ with different $\sigma_{noise}$ to the fine-tuned weights, and find that adding noise to adapter-tuned models leads to much less accuracy degradation than fully fine-tuned models. Adapters still retain most of the performance even if the noise has equivalent variance to the weights (\ie, $\sigma_{noise}=\sigma_{weight}$). Since numerical error can be also viewed as a type of noise, we conjecture that the adapters would not suffer from lower numerical precision.

\subsection{Trading Precision for Efficiency}

In view of the existence of precision redundancy, a natural idea is to trading the redundant precision for much needed efficiency. 
Previous work on quantization \cite{deepcompression,quantization,dkm} has demonstrated that clustering is a reliable direction for quantization of arbitrary bit-width, so we also adopt a clustering-based quantization strategy for adapters.

As illustrated in Figure~\ref{fig:noise}, the smaller the noise, the less the performance degradation. The object of adapter quantization is to minimize the noise involved, \ie, minimize the quantization error. The $b$-bit quantization process can be viewed as dividing $\mathbb{R}$ into $B=2^b$ non-overlapping sets $\left\{\mathcal{U}_1,...,\mathcal{U}_{B}\right\}$, which correspond to a codebook with $B$ codes $\{c_1,...,c_{B}\}$. The quantization function quantizes all values in $\mathcal{U}_{j}$ to $c_j$,
\begin{equation}
    \mathcal{Q}(w) = c_j\ \text{if}\ w\in  \mathcal{U}_j\end{equation}
Then we minimize the quantization error as follows,
\begin{equation}
\begin{aligned}
& \underset{c_1,...,c_{B},\mathcal{U}_1,...,\mathcal{U}_{B}}{\text{minimize}}
& & \sum_{i=1}^m{|w_i - \mathcal{Q}(w_i)|}^p
\end{aligned}
\label{eq4}
\end{equation}
in which $w_i$ is an element of a weight $\boldsymbol{W}$ of the adapters, and $m$ is the number of elements in $\boldsymbol{W}$. This problem is equivalent to 1D clustering, which can be addressed via clustering algorithm such as $k$-means ($p=2$) and $k$-medians ($p=1$).

Low-bit quantization, particularly 1-bit quantization suffers catastrophically  poor performance in the absence of quantization-aware training (QAT). In QAT, the weights are ever-changing, so the clustering algorithm has to be rerun in each forward propagation during tuning. An appropriate clustering algorithm is supposed to have negligible computational cost, but an iterative algorithm like $k$-means and $k$-medians is not efficient enough. Moreover, since the cluster assignment in $k$-means and $k$-medians is not differentiable, this process cannot be end-to-end optimized in QAT. Therefore, although previous work~\cite{deepcompression} has applied $k$-means into post-training quantization, it is not a suitable choice for QAT on adapters.

To find an efficient and differentiable clustering method, we visualize the frequency histogram of the parameters in the weights of adapters. As shown in Figure~\ref{fig:histogram}, we find that the parameters in full-precision adapters are subject to a bell-shaped distribution with tails. For simplicity, we suppose the parameters of each weight are always Gaussian, so that the clustering algorithm can be simplified considerably.

Before tuning, we perform clustering on a standard Gaussian distribution to calculate $\{c_1,...,c_{B}\}$ and $\left\{\mathcal{U}_1,...,\mathcal{U}_{B}\right\}$. 
We suppose $p=1$ in Eq. (\ref{eq4}) and use $k$-medians for simplicity. As illustrated in Figure~\ref{fig:quantization}, in each training step, we first standardize the weights by the means and variances of their parameters,
\begin{equation}w_i' =\frac{w_i-\mu}{\sigma}\end{equation}
where $\mu=\textit{MEAN}(\{w_i\}_{i=1}^m), \sigma=\textit{STD}(\{w_i\}_{i=1}^m)$. According to the Gaussian assumption, the parameters in each standardized weight are subject to standard Gaussian distribution. Then we quantize each standardized weight with the pre-calculated $\{c_1,...,c_{B}\}$ and $\{\mathcal{U}_1,...,\mathcal{U}_{B}\}$,
    \begin{equation}
\hat{w_i}' = \mathcal{Q}(w_i') = c_j\ \text{if}\ w_i'\in  \mathcal{U}_j\end{equation}
Finally, we de-standardize the weights to their original means and variances,
    \begin{equation}
        \hat{w_i} =\hat{w_i}' \cdot \sigma + \mu\end{equation}
and then feed the inputs to perform the forward and backward propagation.

In the whole quantization process, only the quantization operation $\mathcal{Q}$ is not differentiable, so we use straight-through estimator (STE) to approximate the gradient, \ie, $\frac{\partial \mathcal{Q}(w_i')}{\partial w_i'}=1$. Then $\forall w_i,w_k\in\boldsymbol{W}$ the overall gradient is calculated as
\begin{equation}
    \frac{\partial \hat{w_i}}{\partial w_k}=\begin{cases}1+\frac{w_i'(\hat{w_i}' - w_i')}{m}\ \quad\text{if}\ i = k\\\frac{w_k'(\hat{w_i}' - w_i')}{m}\quad\quad\quad\text{otherwise}\end{cases}\end{equation}

During tuning, the pre-trained weights are always frozen, and only the adapters as well as the classification head are updated. The full-precision weights are maintained in training, and updated via end-to-end gradient descent. Since PET only focuses on boosting parameter efficiency, we still use full-precision activation for better performance. After tuning, we store necessary information for reproducing the quantized weights instead of the full-precision adapters, \ie, the $b$-bit quantization indexes $j$ of adapters' parameters ($b$ bits per parameter) and the mean $\mu$ and standard deviation $\sigma$ of each full-precision weight matrix in adapters (128 bits per adapter). $\{c_1,...,c_{B}\}$ and $\{\mathcal{U}_1,...,\mathcal{U}_{B}\}$ can be recalculated before inference. At inference time, the weights are reconstructed as
\begin{equation}
   \hat{w_i}=c_j\cdot\sigma+\mu\end{equation}
which are directly used for inference.

\begin{table}[t]
\begin{center}

\scalebox{0.8}{
\begin{tabular}{p{2.4cm}<{}p{1.8cm}<{}p{2.5cm}<{}p{1.8cm}<{}}

\toprule
Method&Bit-width&Avg. Acc.&Size (KB)\\\hline
\specialrule{0em}{1pt}{1pt}

\multirow{6}{*}{{\sc AdaptFormer}} &\em 32 (FP)&76.70&576.0\\
&\em 16 (FP)&76.70 {(- 0.00)}&288.0\\
&\em 8 (INT)&76.69 \textcolor{red}{($\downarrow$ 0.01)}&144.0\\\cline{2-4}\specialrule{0em}{1pt}{1pt}
&4&76.76 \textcolor{ForestGreen}{($\uparrow$ 0.06)}&72.3\\
&2&76.64 \textcolor{red}{($\downarrow$ 0.06)}&36.2\\
&1&76.41 \textcolor{red}{($\downarrow$ 0.29)}&18.2\\\hline\specialrule{0em}{1pt}{1pt}

\multirow{6}{*}{\sc {\sc LoRA}}&\em 32 (FP)&76.42&1152.0\\
&\em 16 (FP)&76.42 {(- 0.00)}&576.0\\
&\em 8 (INT)&76.42 {(- 0.00)}&288.0\\\cline{2-4}\specialrule{0em}{1pt}{1pt}
&4&76.33 \textcolor{red}{($\downarrow$ 0.09)}&144.4\\
&2&76.27 \textcolor{red}{($\downarrow$ 0.15)}&72.4\\
&1&76.40 \textcolor{red}{($\downarrow$ 0.02)}&36.4\\
\bottomrule
\end{tabular}
}\end{center}
\caption{\textbf{Average accuracy on VTAB-1K benchmark.} We fix $h=8$ for {\sc AdaptFormer} and {\sc LoRA} to 8 and change the bit-width. ``Size'' denotes the size of adapters per task.}
\label{tab:lp1}
\vspace{0pt}
\end{table}

\begin{table}[t]
\begin{center}
\scalebox{0.8}{
\begin{tabular}{p{2.6cm}<{}p{1.8cm}<{}p{1.8cm}<{}p{2.2cm}<{}}

\toprule
Method&Bit-width&Dimension&Avg. Acc.\\\hline
\specialrule{0em}{1pt}{1pt}

\multirow{5}{*}{{\sc AdaptFormer}} &\em 32 (FP)&1&75.29\\
&\em 8 (INT)&4&76.34 \textcolor{ForestGreen}{($\uparrow$ 1.05)}\\\cline{2-4}\specialrule{0em}{1pt}{1pt}
&4&8&76.76 \textcolor{ForestGreen}{($\uparrow$ 1.47)}\\
&2&16&76.89 \textcolor{ForestGreen}{($\uparrow$ 1.60)}\\
&1&32&\textbf{76.97} \textcolor{ForestGreen}{($\uparrow$ 1.68)}\\\hline\specialrule{0em}{1pt}{1pt}
\multirow{5}{*}{{\sc LoRA}}
&\em 32 (FP)&1&75.70\\
&\em 8 (INT)&4&76.08 \textcolor{ForestGreen}{($\uparrow$ 0.38)}\\\cline{2-4}\specialrule{0em}{1pt}{1pt}
&4&8&76.33 \textcolor{ForestGreen}{($\uparrow$ 0.63)}\\
&2&16&{76.70} \textcolor{ForestGreen}{($\uparrow$ 1.00)}\\
&1&32&\textbf{76.72} \textcolor{ForestGreen}{($\uparrow$ 1.02)}\\
\bottomrule
\end{tabular}
}\end{center}
\caption{\textbf{Average accuracy on VTAB-1K benchmark under certain storage budget.} Lower bit-width and higher hidden dimension lead to better performance.}
\label{tab:lp2}
\vspace{0pt}
\end{table}

\begin{table*}[t]

\begin{center}
\setlength{\tabcolsep}{0.3pt}
\scalebox{0.85}{
\begin{tabular}{p{3cm}<{}p{1.25cm}<{\centering}p{0.75cm}<{\centering}|p{0.75cm}<{\centering}p{0.75cm}<{\centering}p{0.75cm}<{\centering}p{0.75cm}<{\centering}p{0.75cm}<{\centering}p{0.75cm}<{\centering}p{0.75cm}<{\centering}|p{0.75cm}<{\centering}p{0.75cm}<{\centering}p{0.75cm}<{\centering}p{0.75cm}<{\centering}|p{0.75cm}<{\centering}p{0.75cm}<{\centering}p{0.75cm}<{\centering}p{0.75cm}<{\centering}p{0.75cm}<{\centering}p{0.75cm}<{\centering}p{0.75cm}<{\centering}p{0.75cm}<{\centering}}
\toprule
\multicolumn{3}{c|}{}&\multicolumn{7}{c|}{\textbf{Natural}}&\multicolumn{4}{c|}{\textbf{Specialized}}&\multicolumn{8}{c}{\textbf{Structured}}\\
&\multicolumn{1}{c}{\STAB{\rotatebox[origin=c]{90}{Size (MB)}}}
&\multicolumn{1}{c|}{\STAB{\rotatebox[origin=c]{90}{Avg. Acc.}}}
&\multicolumn{1}{c}{\STAB{\rotatebox[origin=c]{90}{Cifar100}}}
&\multicolumn{1}{c}{\STAB{\rotatebox[origin=c]{90}{Caltech101}}}
&\multicolumn{1}{c}{\STAB{\rotatebox[origin=c]{90}{DTD}}}
&\multicolumn{1}{c}{\STAB{\rotatebox[origin=c]{90}{Flower102}}}
&\multicolumn{1}{c}{\STAB{\rotatebox[origin=c]{90}{Pets}}}
&\multicolumn{1}{c}{\STAB{\rotatebox[origin=c]{90}{SVHN}}}
&\multicolumn{1}{c|}{\STAB{\rotatebox[origin=c]{90}{Sun397}}}
&\multicolumn{1}{c}{\STAB{\rotatebox[origin=c]{90}{Camelyon}}}
&\multicolumn{1}{c}{\STAB{\rotatebox[origin=c]{90}{EuroSAT}}}
&\multicolumn{1}{c}{\STAB{\rotatebox[origin=c]{90}{Resisc45}}}
&\multicolumn{1}{c|}{\STAB{\rotatebox[origin=c]{90}{Retinopathy}}}
&\multicolumn{1}{c}{\STAB{\rotatebox[origin=c]{90}{Clevr-Count}}}
&\multicolumn{1}{c}{\STAB{\rotatebox[origin=c]{90}{Clevr-Dist}}}
&\multicolumn{1}{c}{\STAB{\rotatebox[origin=c]{90}{DMLab}}}
&\multicolumn{1}{c}{\STAB{\rotatebox[origin=c]{90}{KITTI-Dist}}}
&\multicolumn{1}{c}{\STAB{\rotatebox[origin=c]{90}{dSpr-Loc}}}
&\multicolumn{1}{c}{\STAB{\rotatebox[origin=c]{90}{dSpr-Ori}}}
&\multicolumn{1}{c}{\STAB{\rotatebox[origin=c]{90}{sNORB-Azim}}}
&\multicolumn{1}{c}{\STAB{\rotatebox[origin=c]{90}{sNORB-Ele}}}\\
\specialrule{0em}{1pt}{1pt}
\hline
\specialrule{0em}{1pt}{1pt}
\multicolumn{22}{l}{\emph{Conventional Fine-Tuning}}\\
\hline
\specialrule{0em}{1pt}{1pt}
\sc Full&327&68.9&68.9&87.7&64.3&97.2&86.9&87.4&38.8&79.7&95.7&84.2&73.9&56.3&58.6&41.7&65.5&57.5&46.7&25.7&29.1 \\
\sc Linear&0&57.6&64.4&85.0&63.2&97.0&86.3&36.6&51.0&78.5&87.5&68.5&74.0&34.3&30.6&33.2&55.4&12.5&20.0&9.6&19.2\\
\hline
\specialrule{0em}{1pt}{1pt}
\multicolumn{22}{l}{\emph{PET methods}}\\
\hline
\specialrule{0em}{1pt}{1pt}
{\sc VPT-Deep}~\cite{vpt}&2.03&72.0&\bf78.8&90.8&65.8&98.0&88.3&78.1&49.6&81.8&\bf96.1&83.4&68.4&68.5&60.0&46.5&72.8&73.6&47.9&32.9&37.8 \\
\text{\color{gray}{\sc NOAH}$^\dag$}~\cite{noah}&\text{\color{gray}1.37}&\text{\color{gray}75.5}&\text{\color{gray}69.6}&\text{\color{gray}\bf92.7}&\text{\color{gray}70.2}&\text{\color{gray}99.1}&\text{\color{gray}90.4}&\text{\color{gray}86.1}&\text{\color{gray}53.7}&\text{\color{gray}84.4}&\text{\color{gray}95.4}&\text{\color{gray}83.9}&\text{\color{gray}\bf75.8}&\text{\color{gray}82.8}&\text{\color{gray}\bf68.9}&\text{\color{gray}49.9}&\text{\color{gray}\bf81.7}&\text{\color{gray}81.8}&\text{\color{gray}48.3}&\text{\color{gray}32.8}&\text{\color{gray}44.2}\\
{\sc LoRA}~\cite{lora}&1.13&76.4&72.0&91.2&71.6&99.1&91.3&88.9&56.4&87.2&94.6&83.9&74.9&\bf83.7&64.0&52.3&81.2&84.8&53.3&\bf38.1&43.4\\
{\sc SSF}~\cite{ssf}&0.78&75.7&69.0&92.6&\bf75.1&\bf99.4&\bf91.8&\bf90.2&52.9&87.4&95.9&\bf87.4&75.5&75.9&62.3&\bf53.3&80.6&77.3&54.9&29.5&37.9\\
{\sc Adapter-P}~\cite{adapterp}&0.56&75.5&73.2&90.1&69.6&99.2&91.1&84.9&56.0&86.6&94.8&82.5&\bf75.8&82.9&63.9&49.7&79.7&81.7&55.5&31.6&42.2 \\
{\sc AdaptFormer}~\cite{adaptformer}&0.56&\bf76.7&73.8&92.3&72.7&99.3&91.6&89.1&56.5&\bf87.8&95.5&84.9&75.2&83.3&62.5&52.4&\bf81.7&86.2&\bf55.9&34.4&40.2\\
{\sc BitFit}~\cite{bitfit}&0.39&65.2&72.8&87.0&59.2&97.5&85.3&59.9&51.4&78.7&91.6&72.9&69.8&61.5&55.6&32.4&55.9&66.6&40.0&15.7&25.1\\
{\sc FacT-TT}~\cite{fact}&0.30&\bf76.7&73.4&91.0&72.4&99.2&91.4&90.1&\bf56.6&87.3&94.7&84.5&\bf75.8&83.0&64.9&51.3&81.4&\bf87.4&53.2&33.5&\bf44.3\\
{\sc VPT-Shallow}~\cite{vpt}&0.24&67.8&77.7&86.9&62.6&97.5&87.3&74.5&51.2&78.2&92.0&75.6&72.9&50.5&58.6&40.5&67.1&68.7&36.1&20.2&34.1\\
{\sc Compacter}~\cite{compactor}&\bf0.15&74.2&71.9&89.0&69.7&99.1&90.7&82.7&56.1&86.0&93.5&82.4&75.3&80.2&63.4&47.4&77.2&78.1&53.5&27.3&39.8\\
\hline
\specialrule{0em}{1pt}{1pt}
\rowcolor{lightgray}\multicolumn{2}{l}{{\sc Bi-{\sc LoRA}} (Ours)}&&&&&&&&&&&&&&&&&&&&\\
\rowcolor{lightgray}\quad\ $h=32$&0.14&76.7&72.1&91.7&71.2&99.1&91.4&\bf90.2&55.8&87.0&\bf95.4&85.5&75.5&83.1&64.1&52.2&81.3&\bf86.4&53.5&\bf36.7&\bf44.4\\
\rowcolor{lightgray}\quad\ $h=1$&0.0048&75.4&72.6&90.4&71.8&99.0&91.3&87.0&56.0&86.1&94.1&82.1&75.4&81.0&\bf64.2&50.5&79.7&83.0&53.7&29.7&42.9\\
\rowcolor{lightgray}\multicolumn{2}{l}{{\sc Bi-{\sc AdaptFormer}} (Ours)}&&&&&&&&&&&&&&&&&&&&\\
\rowcolor{lightgray}\quad\ $h=32$&0.071&\bf77.0&\bf74.1&\bf92.4&\bf72.1&\bf99.3&\bf91.6&89.0&\bf56.3&\bf88.2&95.2&\bf86.0&\bf76.2&\bf83.9&63.6&\bf53.0&\bf81.4&86.2&\bf54.8&35.2&41.3\\
\rowcolor{lightgray}\quad\ $h=1$&\bf0.0024&75.0&73.3&91.0&\bf72.1&99.1&91.4&86.0&56.2&87.0&94.6&82.9&76.0&79.6&62.8&50.1&78.6&76.6&53.9&27.4&38.6\\
\bottomrule
\end{tabular}
}
\end{center}
\caption{\textbf{Full results on the VTAB-1K benchmark}. ``Avg. Acc.'' denotes the average results over three groups. ``Size'' denotes the average size of trainable parameters in backbones per task, \ie, classification heads (0.14 MB/task in average) are not counted. $^\dag$~denotes results from~\cite{noah} using normalized inputs.}
\label{tab:vtab}
\vspace{0pt}
\end{table*}

\section{Experiments}
\subsection{Datasets}
We use more than 20 image classification tasks to evaluate the performance of different PET methods.

\textbf{VTAB-1K benchmark. }VTAB-1K~\cite{vtab} contains 19 image classification tasks from diverse fields, which can be categorized into three groups: Natural, Specialized, and Structured. These tasks cover a large range of possible domains where downstream tasks come, so the performance of different methods on this benchmark largely reflects their ability to transfer learning. Each dataset contains 800 samples for training and 200 for validation. Following previous work~\cite{vpt,noah,ssf,fact,convpass}, we tune the pre-trained model with all the 1,000 training and validation samples and report results evaluated on test-set. Following~\cite{vpt,ssf}, we use \emph{unnormalized inputs} that are consistent with the VTAB paper~\cite{vtab}. Note that some previous methods~\cite{noah,fact} normalize the images with ImageNet's mean and standard deviation, so we re-implement some of them for a fair comparison.

\textbf{Few-shot fine-grained visual recognition (FGVC). }We use five FGVC datasets to evaluate the capability of PET methods in the low-data regime. The five datasets are {FGVC-Aircraft}~\cite{aircraft}, {Oxford-Pets}~\cite{pets}, {Food-101}~\cite{food}, {Stanford Cars}~\cite{car}, and {Oxford-Flowers102}~\cite{flower}. Experiments are conducted in 1, 2, 4, 8, and 16-shot settings.


\subsection{Performance of Low-Precision Adapters}
We first address the most critical question in this paper: \emph{is reducing precision redundancy a good choice for improving the parameter efficiency of adapter-based PET methods?} To investigate the role of numerical precision in adapters, we make comparisons across different bit-widths. We use {\sc AdaptFormer}~\cite{adaptformer} and {\sc LoRA}~\cite{lora} with $h=8$ to adapt ViT-B/16~\cite{vit} pre-trained on supervised ImageNet-21K~\cite{imagenet}. The 32-bit adapters are trained using FP32 without quantization. 16-bit (FP16) and 8-bit (INT8) adapters are directly converted from fine-tuned FP32 adapters. Others are fine-tuned using the proposed QAT method. Table~\ref{tab:lp1} presents the accuracy and adapter size on VTAB-1K.

We notice that using $b$-bit adapters leads to about $\frac{32}{b}\times$ more parameter efficiency than full-precision adapters. However, the performance degradation resulting from quantization is very slight and sometimes negligible, even in the 1-bit setting.
Note that quantizing the entire model to a very low bit-width usually causes significant performance degradation, but our observation indicates that low-bit quantization only on adapters is reliable and much less damaging.

Moreover, we explore the best bit-width given a certain storage budget. Since low-precision adapters are more lightweight, we can augment their performance by using higher hidden dimension to utilize the saved space. The size of a $b$-bit $h$-dimension adapter is about $2dbh$ bits where $d$ is the feature dimension, so we fix $bh=32$ and compare different combinations of $b$ and $h$. As shown in Table~\ref{tab:lp2}, the lower $b$ and higher $h$ yield better performance on {\sc LoRA} and {\sc AdaptFormer}. 1-bit adapters perform the best across different combinations. Overall, we find that the parameter efficiency gains of the low-bit adapters far outweigh their performance damage, demonstrating the feasibility and necessity to trade precision for efficiency.

\begin{figure*}[t]
     \centering
         \includegraphics[width=0.95\textwidth]{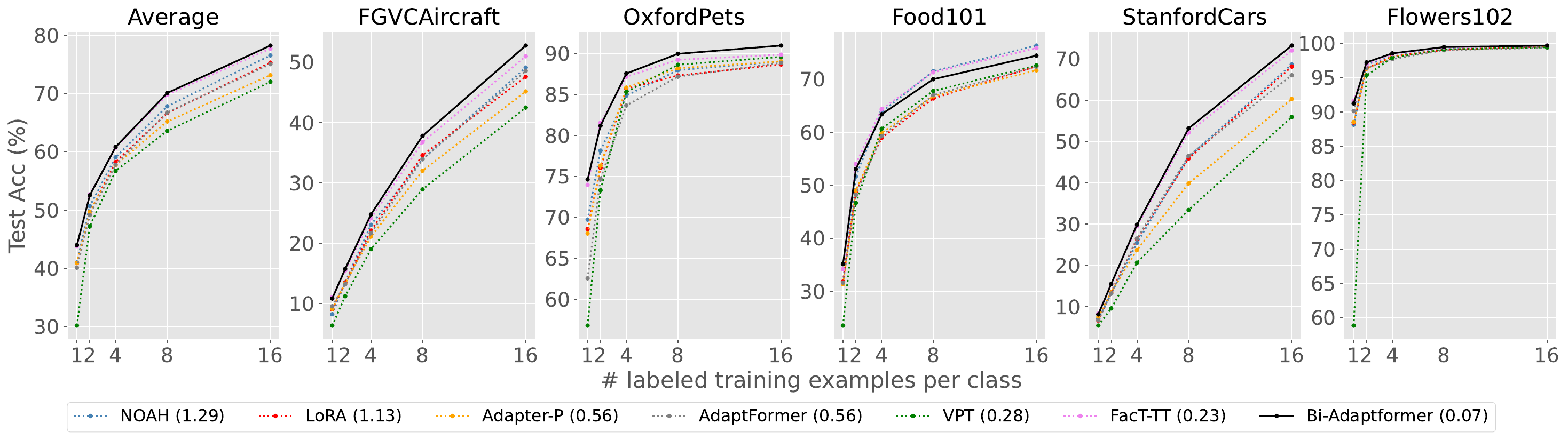}
         \caption{\textbf{Accuracy of few-shot learning on FGVC datasets.} The average size (MB) of trainable parameters in backbones is shown in parentheses. {\sc Bi-{\sc AdaptFormer}} outperforms other baselines on average accuracy using the fewest trainable parameters. Results are averaged over three trials with different seeds.}
         \label{fig:fs}
         \vspace{0pt}
\end{figure*}

\begin{table*}[t]
\begin{center}
 \begin{subtable}[h]{0.51\textwidth}
\begin{center}
\setlength{\tabcolsep}{0.3pt}
\scalebox{0.8}{
\begin{tabular}{p{3cm}<{}p{1cm}<{}p{2.2cm}<{}p{2.4cm}<{}p{2.2cm}<{}}
\toprule

Method&$h$&Binary Head&Avg. Acc.&{Ckpt Size (KB)}\\\hline
\specialrule{0em}{1pt}{1pt}
\sc Full&-&$\times$&68.9&3.4$\times$10$^5$\\
\sc Linear&-&$\times$&57.6&140.8\\\hline\specialrule{0em}{1pt}{1pt}
\multirow{4}{*}{{\STAB{\rotatebox[origin=c]{0}{\makecell[c]{{\sc Bi-{\sc AdaptFormer}}}}}}}&32&$\times$&76.97&212.9\\
&32&$\checkmark$&76.89 \textcolor{red}{($\downarrow$ 0.08)}&76.6\\\cline{2-5}\specialrule{0em}{1pt}{1pt}
&1&$\times$&74.96&143.2\\
&1&$\checkmark$&73.81 \textcolor{red}{($\downarrow$ 1.15)}&\bf 6.8\\\hline\specialrule{0em}{1pt}{1pt}
\multirow{4}{*}{{\STAB{\rotatebox[origin=c]{0}{\makecell[c]{{\sc Bi-{\sc LoRA}}}}}}}&32&$\times$&76.72&285.1\\
&32&$\checkmark$&76.31 \textcolor{red}{($\downarrow$ 0.41)}&148.8\\\cline{2-5}\specialrule{0em}{1pt}{1pt}
&1&$\times$&75.39&145.6\\
&1&$\checkmark$&74.56 \textcolor{red}{($\downarrow$ 0.83)}&\bf 9.2\\
\bottomrule

\end{tabular}
}\end{center}
\vspace{-8pt}
\caption{\textbf{Classification head quantization.} ``Ckpt Size'' denotes the average size of checkpoint including classification heads.}
\label{tab:head}
\end{subtable}
\hfill
 \begin{subtable}[h]{0.45\textwidth}
 \begin{center}
\setlength{\tabcolsep}{0.3pt}
\scalebox{0.8}{
\begin{tabular}{p{2.5cm}<{}p{3cm}<{}p{2.5cm}<{}p{1.8cm}<{}}
\toprule

Backbone&Method&Avg. Acc.&Size (MB)\\\hline
\specialrule{0em}{1pt}{1pt}
\multirow{5}{*}{ConvNeXt-B}&\sc Full&74.03&334.0\\
&\sc Linear&63.58&0\\

&{\sc VPT-Deep}&68.71&0.017\\
&{\sc AdaptFormer}&78.86&1.102\\
\rowcolor{lightgray}\cellcolor{white}&{{\sc Bi-{\sc AdaptFormer}}}&\textbf{79.07} \textcolor{ForestGreen}{($\uparrow$ 0.21)}&0.138\\\hline\specialrule{0em}{1pt}{1pt}
\multirow{5}{*}{Swin-B}&Full&74.99&332.2\\
&Linear&62.60&0\\

&{\sc VPT-Deep}&71.55&0.622\\
&{\sc AdaptFormer}&77.22&0.734\\

\rowcolor{lightgray}\cellcolor{white}&{{{\sc Bi-{\sc AdaptFormer}}}}&\textbf{77.24} \textcolor{ForestGreen}{($\uparrow$ 0.02)}&0.092\\
\bottomrule
\end{tabular}
}\end{center}
\vspace{-8pt}
\caption{\textbf{Performance on other backbones.} We use ConvNeXt-B and Swin-B as backbones.}
\label{tab:backbone}

\end{subtable}
\hfill
\vspace{10pt}
 \begin{subtable}[h]{0.34\textwidth}
 \setlength{\tabcolsep}{0.3pt}

 \begin{center}
\scalebox{0.75}{
\begin{tabular}{p{3cm}<{}p{2.4cm}<{}p{2.2cm}<{}}

\toprule

Method&\makecell[l]{Training Time\\(ms/batch)}&\makecell[l]{Inference Time\\(ms/batch)}\\\hline
\specialrule{0em}{1pt}{1pt}
\sc Full&342.0&\bf110.8\\
\sc Linear&\bf115.2&\bf110.8\\
{\sc VPT-Deep}&407.9&174.2\\
{\sc FacT-TT}&296.0&\bf110.8\\\hline\specialrule{0em}{1pt}{1pt}
{\sc AdaptFormer}&252.3&115.3\\
\rowcolor{lightgray}{\sc Bi-{\sc AdaptFormer}}&265.0 \textcolor{red}{($\uparrow$ 12.7)}&115.5 \textcolor{red}{($\uparrow$ 0.02)}\\\hline\specialrule{0em}{1pt}{1pt}
{\sc LoRA}&275.7&\bf110.8\\
\rowcolor{lightgray}{\sc Bi-{\sc LoRA}}&293.2 \textcolor{red}{($\uparrow$ 17.5)}&\bf110.8\\
\bottomrule
\end{tabular}
}\end{center}
\vspace{-8pt}
\caption{\textbf{Average training and inference time.} Measured on a single GeForce RTX
3090 GPU with batch size 64.}
\label{tab:time}

\end{subtable}
\iftrue
\hfill
 \begin{subtable}[h]{0.3\textwidth}
 \begin{center}
  \setlength{\tabcolsep}{0.3pt}
\scalebox{0.75}{
\begin{tabular}{p{2.6cm}<{}p{1.6cm}<{}p{2.2cm}<{}}

\toprule
Method&Bit-width&Avg. Acc.\\\hline
\specialrule{0em}{1pt}{1pt}

\multirow{4}{*}{{\sc AdaptFormer}}
&2 (PTQ) &74.89 \\
&1 (PTQ) &67.28\\\cline{2-3}\specialrule{0em}{1pt}{1pt}
&\cellcolor{lightgray}2 (Ours) &\cellcolor{lightgray}\textbf{76.64} \textcolor{ForestGreen}{($\uparrow$ 1.75)}\\
&\cellcolor{lightgray}1 (Ours) &\cellcolor{lightgray}\textbf{76.41} \textcolor{ForestGreen}{($\uparrow$ 9.13)}\\\hline\specialrule{0em}{1pt}{1pt}

\multirow{4}{*}{{\sc LoRA}}
&2 (PTQ) &74.22\\
&1 (PTQ) &67.38\\\cline{2-3}\specialrule{0em}{1pt}{1pt}
&\cellcolor{lightgray}2 (Ours) &\cellcolor{lightgray}\textbf{76.27} \textcolor{ForestGreen}{($\uparrow$ 2.05)}\\
&\cellcolor{lightgray}1 (Ours) &\cellcolor{lightgray}\textbf{76.40} \textcolor{ForestGreen}{($\uparrow$ 9.02)}\\
\bottomrule
\end{tabular}
}\end{center}
\vspace{-8pt}
\caption{\textbf{QAT \emph{vs.} PTQ.} ``PTQ'' denotes directly quantizing fine-tuned FP32 adapters using $k$-means.}
\label{tab:ab1}
\end{subtable}
\hfill
 \begin{subtable}[h]{0.34\textwidth}
 \begin{center}
 \setlength{\tabcolsep}{0.3pt}

\scalebox{0.75}{
\begin{tabular}{p{3cm}<{}p{1.4cm}<{}p{1.7cm}<{}p{1.5cm}<{}}

\toprule
Method&\# Block&Avg. Acc.&Size (KB)\\\hline
\specialrule{0em}{1pt}{1pt}

\rowcolor{lightgray}\cellcolor{white}\multirow{3}{*}{{\sc Bi-{\sc AdaptFormer}}}&1&\bf76.97&\bf72.2\\
&8&76.96&73.5\\
&32&76.92&78.0\\\hline\specialrule{0em}{1pt}{1pt}

\rowcolor{lightgray}\cellcolor{white}\multirow{3}{*}{{\sc Bi-{\sc LoRA}}}&1&\bf76.72&\bf144.4\\
&8&76.69&147.0\\
&32&76.66&156.0\\
\bottomrule
\end{tabular}

}\end{center}
\vspace{-8pt}
\caption{\textbf{Block-wise quantization.} We use {\sc Bi-{\sc AdaptFormer}} and {\sc Bi-{\sc LoRA}} with $h=32$.}
\label{tab:ab2}
\end{subtable}
\fi
\end{center}
\vspace{-6pt}
\caption{\textbf{Supplementary results on VTAB-1K benchmark.}}

\end{table*}
\vspace{-8pt}
\subsection{Comparison with the State-of-the-Art}

\subsubsection{VTAB-1K benchmark}
We compare our methods with full fine-tuning, linear probing (\ie, only training the classification head), {\sc VPT}~\cite{vpt}, {\sc NOAH}~\cite{noah}, {\sc SSF}~\cite{ssf}, {\sc Adapter-P}~\cite{adapterp}, {\sc BitFit}~\cite{bitfit}, {\sc AdaptFormer}~\cite{adaptformer}, {\sc LoRA}~\cite{lora}, {\sc Compacter}~\cite{compactor}, and {\sc FacT}~\cite{fact} on VTAB-1K. All baselines use FP32 by default. The hidden dimension $h$ is set to 8 for {\sc Adapter-P}, {\sc AdaptFormer}, and {\sc LoRA}. The number of Kronecker products and hidden dimensions are 4 and 32 for {\sc Compacter}, respectively. For {\sc FacT}, we use {\sc FacT-TT} with rank searched from \{8, 16, 32\} to adapt the MHSA blocks. The settings of other baselines follow their original papers. As for our low-precision adapters, we quantize the bit-width of {\sc AdaptFormer} and {\sc LoRA} to 1, named \textbf{{\sc Bi-{\sc AdaptFormer}}} and \textbf{{\sc Bi-{\sc LoRA}}}, and report results with hidden dimensions $h=1$ and $32$. All these methods use a ViT-B/16~\cite{vit} pre-trained on supervised ImangeNet-21K as backbone. We train the models for 100 epochs with AdamW optimizer. 

Table~\ref{tab:vtab} shows the full results on VTAB-1K. Since 1-bit adapters are much more storage-efficient than their full-precision counterparts, {\sc Bi-{\sc AdaptFormer}} and {\sc Bi-{\sc LoRA}} can use a larger hidden dimension while maintaining a smaller size. Our {\sc Bi-{\sc AdaptFormer}} with $h=32$ beats all previous PET methods while using a smaller storage size. Notably, {\sc Bi-{\sc AdaptFormer}} and {\sc Bi-{\sc LoRA}} achieve better performance than {\sc Compacter} and {\sc FacT-TT} while being more parameter-efficient, indicating that precision redundancy is more significant than rank redundancy in adapters and thus quantization is a better solution than low-rank parameterization for designing efficient adapters. %
Moreover, {\sc Bi-{\sc AdaptFormer}} and {\sc Bi-{\sc LoRA}} with $h=1$ only store less than 5 KB of backbone parameters for each task, while reaching performance better than {\sc VPT}, {\sc BitFit}, {\sc Compacter}, and full fine-tuning.

\subsubsection{Few-shot learning on FGVC}
On few-shot FGVC datasets, we compare {\sc Bi-{\sc AdaptFormer}}, the best-performing quantized adapter in the experiments above, with other competitive baselines: {\sc VPT-Deep}, {\sc Adapter-P}, {\sc LoRA}, {\sc AdaptFormer}, {\sc NOAH}, and {\sc FacT-TT}. The hidden dimensions of {\sc Adapter-P}, {\sc LoRA}, and {\sc AdaptFormer}, as well as the prompt length of {\sc VPT-Deep}, are all set to 8. The rank of {\sc FacT-TT} is set to 16, and {\sc NOAH} follows the best recipes in~\cite{noah}. As for {\sc Bi-{\sc AdaptFormer}}, we use a hidden dimension of 32. Other settings are the same as in the VTAB-1K experiments.  Per-dataset results as well as the average results in the five settings are shown in Figure~\ref{fig:fs}.

Overall, our {\sc Bi-{\sc AdaptFormer}} outperforms all baselines on 5-task average accuracy with the smallest size of trainable parameters. On FGVC-Aircraft, Oxford-Pets, and Stanford Cars, {\sc Bi-{\sc AdaptFormer}} exhibits significant performance improvement over the previously state-of-the-art PET methods. Only on Food-101, {\sc Bi-{\sc AdaptFormer}} performs worse than {\sc FacT-TT} and {\sc NOAH}. Note that {\sc Bi-{\sc AdaptFormer}} is about 3$\times$ and 19$\times$ more storage-efficient than {\sc FacT} and {\sc NOAH}, respectively, and thus is more competitive under strict storage restrictions.

\subsection{Further Analysis}
\subsubsection{Quantizing classification head}
As the size of adapters is compressed, the classification heads take up most of the storage space, hindering further improvements in storage efficiency. For example, on VTAB-1K, the average size of the classification heads is 0.14 MB, much larger than that of {\sc Bi-{\sc AdaptFormer}} modules.
As shown in Table~\ref{tab:head}, by quantizing the classification heads, {\sc Bi-{\sc AdaptFormer}} keeps state-of-the-art results (76.89 \emph{vs.} {\sc AdaptFormer}'s 76.70) with checkpoint size smaller than linear probing (76.7 KB \emph{vs.} 140.8 KB). Note that linear probing is usually considered as the efficiency lower bound of adaptation. Furthermore, {\sc Bi-{\sc AdaptFormer}} and {\sc Bi-{\sc LoRA}} with $h=1$ and binary head achieve better performance than full fine-tuning, linear probing, and {\sc VPT}, but the average size of the total checkpoints is only 6.8 KB and 9.2 KB, respectively, which are dozens of times more storage-efficient than linear probing. 

\subsubsection{Computational efficiency}

One of the design principles behind our quantization method is to ensure the quantization operation has negligible computational cost during QAT. To evaluate the efficiency of our proposed method, we conducted experiments to study the training and inference time of different tuning methods, as summarized in Table~\ref{tab:time}. For all baselines, we use the same settings as in the VTAB-1K experiments. As for our {\sc Bi-{\sc AdaptFormer}} and {\sc Bi-{\sc LoRA}}, we set a larger hidden dimension $h=32$. We find that the QAT and larger $h$ slightly increase the training time of adapters. 
However, {\sc Bi-{\sc AdaptFormer}} and {\sc Bi-{\sc LoRA}} are still faster than {\sc VPT}, {\sc FacT}, and full fine-tuning. At inference time, since  {\sc (Bi-){\sc LoRA}}, and {\sc FacT} can be re-parameterized and absorbed into the pre-trained backbone, they do not incur additional computation. 

\subsubsection{Performance on other backbones}
Note that our proposed quantization method is a plug-in strategy that can be applied in any backbones and any adapters. Besides ViTs~\cite{vit}, there are also other commonly  used backbone networks in vision, such as hierarchical transformers like Swin~\cite{swin} and convolutional networks like ConvNeXt~\cite{convnext}. In Table~\ref{tab:backbone}, we apply {\sc Bi-{\sc AdaptFormer}} to Swin-B and ConvNeXt-B, and compare it with other baselines that can also be extended to these backbones. We notice that {\sc Bi-{\sc AdaptFormer}} still achieves state-of-the-art results on VTAB-1K. {\sc Bi-{\sc AdaptFormer}} with $h=32$ offers on-par or better performance than {\sc AdaptFormer} with $h=8$ while only using about $\frac{1}{8}$ of the storage size, which verifies the generalization ability of binary adapters.

\subsubsection{Ablation studies}
We perform further ablation experiments on our low-bit adapters. The low-bit adapters are fine-tuned via QAT, which has been proven to work better in low-bit settings. To illustrate this, we compare our method with a PTQ method, \ie, directly quantizing fine-tuned full-precision adapters using $k$-means. We set $h=8$ for {\sc AdaptFormer} and {\sc LoRA}. As shown in Table \ref{tab:ab1}, PTQ obviously underperforms QAT, especially in 1-bit setting.

Moreover, since each weight matrix can be divided into several sub-matrices as blocks to perform block-wise quantization, \ie, standardizing the parameters and storing the $\mu$ and $\sigma$ of each block, we here compare the performance of 1-bit adapters across different numbers of blocks. We set $h=32$ for all methods. As shown in Table \ref{tab:ab2}, since block-wise quantization methods  (\# block $>1$) store more $\mu$ and $\sigma$ than our methods (\# block $=1$), block-wise quantization uses a larger storage size. However, block-wise quantization does not demonstrate superiority over our methods.

\section{Conclusion}
In this work, we systematically revisit the parameter efficiency of adapter-based PET through the lens of precision redundancy. Based on our observations, we propose a plug-in strategy to train low-precision counterparts for existing adapter-based methods. Through extensive experiments on more than 20 datasets, we empirically verify the superiority of 1-bit adapters in terms of both performance and parameter efficiency. Surprisingly, we find that 2.4 KB parameters in backbone 
is almost sufficient to describe the difference between the pre-trained ViT-B and a task-specific fine-tuned ViT-B, suggesting that the intrinsic dimension of visual datasets is much smaller than what we used to believe. Our work also brings quantization to PET, providing a general solution to largely enhance the parameter efficiency of adapter-based PET methods.

{\small
\bibliographystyle{ieee_fullname}
\bibliography{egbib}}

\clearpage

\clearpage
\section*{Appendix}
\setcounter{section}{0}
\section{Algorithm}
\begin{algorithm}[h]
  \SetAlgoLined
  \KwIn{Bit-width $b$}
  \KwOut{Sets $\mathcal{U}_1,...,\mathcal{U}_{2^b}$ and codes $c_1,...,c_{2^b}$}

  Initialize all $c_i$ to 0\;
  Set $c_0=-\infty,c_{2^b+1}=+\infty$\;
  \While{not converged}{
    $\forall i, \mathcal{U}_{i}=(\frac{c_{i-1}+c_i}{2}, \frac{c_i+c_{i+1}}{2}]$\;
    $\forall i, c_i = $ median of standard Gaussian within $\mathcal{U}_i$\;
  }

  \Return{$\mathcal{U}_1,...,\mathcal{U}_{2^b}, c_1,...,c_{2^b}$}\;

  \caption{$k$-medians on Standard Gaussian}
  \label{alg}

\end{algorithm}

\section{Derivation of Eq. (8)}
If $i=k$,
\begin{align*}
    \frac{\partial \hat{w_i}}{\partial w_k}&=   \frac{\partial (\hat{w_i}'\cdot \sigma + \mu)}{\partial w_i}\\
    &=\sigma\cdot\frac{\partial \hat{w_i}'}{\partial w_i}+\hat{w_i}'\cdot\frac{\partial \sigma}{\partial w_i}+\frac{\partial \mu}{\partial w_i}\\    &=\sigma\cdot\frac{\partial {w_i}'}{\partial w_i}+\hat{w_i}'\cdot\frac{w_i'}{m}+\frac{1}{m}\\
    &=\sigma\cdot\frac{\partial\frac{w_i-\mu}{\sigma}}{\partial w_i}+\hat{w_i}'\cdot\frac{w_i'}{m}+\frac{1}{m}\\
        &=\sigma\cdot\frac{m-1-w_i'^2}{m\sigma}+\hat{w_i}'\cdot\frac{w_i'}{m}+\frac{1}{m}\\
        &=1+\frac{w_i'(\hat{w_i}' - w_i')}{m}\\
\end{align*}
\quad Otherwise,
\begin{align*}
    \frac{\partial \hat{w_i}}{\partial w_k}&=   \frac{\partial (\hat{w_i}'\cdot \sigma + \mu)}{\partial w_k}\\
    &=\sigma\cdot\frac{\partial \hat{w_i}'}{\partial w_k}+\hat{w_i}'\cdot\frac{\partial \sigma}{\partial w_k}+\frac{\partial \mu}{\partial w_k}\\    &=\sigma\cdot\frac{\partial {w_i}'}{\partial w_k}+\hat{w_i}'\cdot\frac{w_k'}{m}+\frac{1}{m}\\
    &=\sigma\cdot\frac{\partial\frac{w_i-\mu}{\sigma}}{\partial w_k}+\hat{w_i}'\cdot\frac{w_k'}{m}+\frac{1}{m}\\
        &=\sigma\cdot\frac{-1-w_i'w_k'}{m\sigma}+\hat{w_i}'\cdot\frac{w_k'}{m}+\frac{1}{m}\\
        &=\frac{w_k'(\hat{w_i}' - w_i')}{m}\\
\end{align*}

\section{Supplementary Experiments}
\subsection{Performance on Full CIFAR100}
In the VTAB-1K benchmark, each task only contains 1,000 training samples. We also conduct experiments on the full CIFAR100~\cite{krizhevsky2009learning}  dataset, which has a larger 60,000-image training set. Following~\cite{ssf}, we use a ViT-B/16 supervisedly pre-trained on ImageNet-21K with AugReg as backbone, and train the model for 100 epochs with batch size 128. We use $h=8$ for {\sc AdaptFormer} and $h=32$ for {\sc Bi-{\sc AdaptFormer}}. All other settings are the same as in~\cite{ssf}.  As shown in Table~\ref{tab:cifar}, {\sc Bi-{\sc AdaptFormer}} still outperforms other baselines while using the smallest storage size. Compared to {\sc AdaptFormer}, {\sc Bi-{\sc AdaptFormer}} brings 0.4\% performance improvement and 8$\times$ more storage efficiency.
\begin{table}[h]
\begin{center}
\setlength{\tabcolsep}{0.3pt}
\scalebox{0.88}{
\begin{tabular}{p{3.8cm}<{\centering}p{2.0cm}<{\centering}p{2cm}<{\centering}}
\hline\specialrule{0em}{1pt}{1pt}
Method&Top-1 Acc.&Size (MB)\\\hline
\specialrule{0em}{1pt}{1pt}
\sc Full$^\dag$&93.82&334.0\\
\sc Linear$^\dag$&88.70&0\\
{\sc BitFit}$^\dag$&93.39&0.39\\
{\sc VPT-Shallow}$^\dag$&90.38&0.59\\
{\sc VPT-Deep}$^\dag$&93.17&1.76\\
{\sc SSF}$^\dag$&\bf 93.99&0.78\\\hline\specialrule{0em}{1pt}{1pt}
{\sc AdaptFormer}&93.55&0.56\\
\rowcolor{lightgray}{{\sc Bi-{\sc AdaptFormer}}}&\underline{93.95} \textcolor{ForestGreen}{($\uparrow$ 0.40)}&\bf0.071\\
\bottomrule
\end{tabular}
}\end{center}
\caption{\textbf{Accuracy on full CIFAR100.} $^\dag$\ denotes results reported in~\cite{ssf}.}
\label{tab:cifar}

\end{table}
\subsection{Semantic Segmentation}
As for the dense prediction, we apply our method on Segmenter~\cite{seg}. We use DeiT-B/16$_{384}$~\cite{deit} pre-trained on ImageNet-1K as encoder. Since each segmentation task tunes an individual decoder upon the pre-trained encoder, we use a single FC layer as a decoder which is much more lightweight than FCN~\cite{fcn} and MaskTransformer~\cite{seg}. We conduct experiments on Pascal-Context~\cite{pascal}. We evaluate three tuning paradigms: full fine-tuning, Adaptformer with $h=8$, and {\sc Bi-{\sc AdaptFormer}} with $h=32$. The models are trained for 50 epochs with batch size 64. As shown in Table~\ref{tab:seg}, {\sc Bi-{\sc AdaptFormer}} still outperform {\sc AdaptFormer} in terms of both performance and efficiency.
\begin{table}[h]
\begin{center}
\setlength{\tabcolsep}{0.3pt}
\scalebox{0.88}{
\begin{tabular}{p{3.2cm}<{\centering}p{2.0cm}<{\centering}p{2cm}<{\centering}}
\hline
\specialrule{0em}{1pt}{1pt}
Method&mIoU (SS)&Size (MB)\\\hline
\specialrule{0em}{1pt}{1pt}
\sc Full&\bf52.61&335.1\\
{\sc AdaptFormer}&51.57&0.56\\
\rowcolor{lightgray}{{\sc Bi-{\sc AdaptFormer}}}&{51.75} \textcolor{ForestGreen}{($\uparrow$ 0.18)}&\bf0.071\\
\bottomrule
\end{tabular}
}\end{center}
\caption{\textbf{Semantic segmentation on Pascal-Context.} ``Size'' denotes the size of trainable parameters in encoders. We report mIoU of single-scale inference on validation set. Each method also has a decoder of 0.17MB.}
\label{tab:seg}
\end{table}

\subsection{Comparison with Other Quantization Methods}
We compare our quantization method with existing binary neural networks -- XNOR-Net~\cite{xnor} and IR-Net~\cite{irnet}. Similar to {\sc Bi-{\sc AdaptFormer}}, we use the quantization strategy of XNOR-Net and IR-Net to quantize (\ie, binarize) the weights of adapters to 1 bit, and keep the full-precision activation, called {\sc XNOR-{\sc AdaptFormer}} and {\sc IR-{\sc AdaptFormer}}, respectively. In experiments, we find that {\sc IR-{\sc AdaptFormer}} cannot be trained stably without Batch Normalization (BN)~\cite{bn}, so we add BN after each FC layer of the adapters. We set the hidden dimension $h=8$ for {\sc AdaptFormer}, and $h=32$ for {\sc Bi-{\sc AdaptFormer}}, {\sc XNOR-{\sc AdaptFormer}}, and {\sc IR-{\sc AdaptFormer}}.

As shown in Table~\ref{tab:irnet}, since {\sc IR-{\sc AdaptFormer}} uses additional BN and {\sc XNOR-{\sc AdaptFormer}} uses channel-wise scaling factors, their storage sizes are larger than that of {\sc Bi-{\sc AdaptFormer}}. {\sc IR-{\sc AdaptFormer}} results in significant performance degradation compared to {\sc AdaptFormer}. We conjunct this is because IR-Net is designed for traditional convolutional networks equiped with BN and is not suitable for plug-in adapters in modern large-size vision architecture. {\sc XNOR-{\sc AdaptFormer}} also underperforms {\sc AdaptFormer}, demonstrating the necessity of tailoring quantization strategy for adapters.

\begin{table}[h]

\begin{center}

\setlength{\tabcolsep}{0.3pt}
\scalebox{0.88}{
\begin{tabular}{p{3.8cm}<{\centering}p{2.7cm}<{\centering}p{2cm}<{\centering}}
\hline
\specialrule{0em}{1pt}{1pt}
Method&Avg. Acc.&Size (MB)\\\hline
\specialrule{0em}{1pt}{1pt}
\sc Full&68.9&334.0\\
\sc Linear&57.6&0\\
{\sc AdaptFormer}&76.70&0.56\\
{\sc IR-{\sc AdaptFormer}}&72.13 \textcolor{red}{($\downarrow$ 4.57)}&0.14\\
{\sc XNOR-{\sc AdaptFormer}}&76.34 \textcolor{red}{($\downarrow$ 0.36)}&0.11\\
\rowcolor{lightgray}{{\sc Bi-{\sc AdaptFormer}}}&\textbf{76.97} \textcolor{ForestGreen}{($\uparrow$ 0.27)}&\bf0.071\\
\bottomrule
\end{tabular}
}\end{center}
\caption{\textbf{Average accuracy on VTAB-1K.}}
\label{tab:irnet}

\end{table}


\section{Experimental Details}
\subsection{Datasets}
See Table~\ref{tab:dataset}.
\subsection{Pre-Trained Backbones}

\begin{table}[h]

\begin{center}
\setlength{\tabcolsep}{0.3pt}
\scalebox{0.8}{

\begin{tabular}{p{3.8cm}<{\centering}p{2.7cm}<{\centering}p{1.5cm}<{\centering}p{2cm}<{\centering}}
\hline
\specialrule{0em}{1pt}{1pt}
Model&\makecell[c]{Pre-Training\\Dataset}&Size (M)&\makecell[c]{Pre-Trained\\Weights}
\\\hline\specialrule{0em}{1pt}{1pt}
ViT-B/16~\cite{vit}&ImageNet-21K&85.8&\href{https://storage.googleapis.com/vit_models/imagenet21K/ViT-B_16.npz}{checkpoint}
\\
Swin-B~\cite{swin}&ImageNet-21K&86.7&\href{https://github.com/SwinTransformer/storage/releases/download/v1.0.0/swin_base_patch4_window7_224_22k.pth}{checkpoint}
\\
ConvNeXt-B~\cite{convnext}&ImageNet-21K&87.6&\href{https://dl.fbaipublicfiles.com/convnext/convnext_base_22k_224.pth}{checkpoint}\\
AugReg ViT-B/16~\cite{vit}&ImageNet-21K&85.8&\href{https://storage.googleapis.com/vit_models/augreg/B_16-i21k-300ep-lr_0.001-aug_medium1-wd_0.1-do_0.0-sd_0.0.npz}{checkpoint}
\\
DeiT-B/16$_{384}$~\cite{convnext}&ImageNet-1K&86.1&\href{https://dl.fbaipublicfiles.com/deit/deit_base_patch16_384-8de9b5d1.pth}{checkpoint}\\
\bottomrule
\end{tabular}
}
\end{center}
\caption{\textbf{Pre-Trained backbones.}
}
\label{tab:pt}
\end{table}

\subsection{Code Implementation}
We use \href{https://pytorch.org/}{\emph{PyTorch}} and \href{https://rwightman.github.io/pytorch-image-models/}{\emph{timm}} to implement all experiments on NVIDIA RTX 3090 GPUs.

\subsection{Data Augmentation}
\subsubsection{VTAB-1K} Following \cite{vpt}, we just resize the images to $224\times224$.

\subsubsection{Few-shot learning} Following \cite{noah}, for training samples, we use color-jitter and RandAugmentation; for validation/test samples, we resize them to $256\times256$, crop them to $224\times224$  at the center, and then normalize them with ImageNet's mean and standard deviation.
\subsubsection{Full CIFAR100} Following~\cite{ssf}, we use a strong augmentation in the fine-tuning setting of~\cite{vit}. Please refer to the official code of~\cite{vit}.

\subsubsection{Semantic Segmentation} We completely follow the setting used in~\cite{seg}, which does mean substraction, random
resizing, random left-right flipping, and randomly crops large images and pad small images to $480\times480$.

\subsection{Hyper-parameters}
$s$ of {\sc (Bi-){\sc AdaptFormer}}, {\sc (Bi-){\sc LoRA}}, and {\sc FacT} is searched from \{0.01, 0.1, 1, 10, 100\}. See Table~\ref{tab:hyper} for other hyper-parameters. We basically follow the hyper-parameters used by~\cite{noah}.

\begin{table*}[t]

\begin{center}
\scalebox{0.9}{
\begin{tabular}{clcccc}
\toprule[1.5pt]
                          & Dataset              & \# Classes & Train                    & Val   & Test            \\ \midrule
                          \multicolumn{6}{c}{VTAB-1K~\cite{vtab}}\\\midrule
\multirow{7}{*}{Natural} & CIFAR100~\cite{krizhevsky2009learning}             & 100       & \multirow{7}{*}{800/1,000}                 & \multirow{7}{*}{200}   & 10,000           \\
                          & Caltech101~\cite{fei2004learning}           & 102       &                 &    & 6,084            \\
                          & DTD~\cite{cimpoi14describing}                 & 47        &                 &    & 1,880            \\
                          & Oxford-Flowers102~\cite{flower}    & 102       &                 &    & 6,149            \\
                          & Oxford-Pets~\cite{pets}          & 37        &                 &    & 3,669     \\
                          & SVHN~\cite{netzer2011reading}                 & 10        &                 &    & 26,032            \\
                          & Sun397~\cite{xiao2010sun}               & 397       &                 &    & 21,750           \\
\midrule\multirow{4}{*}{Specialized}                          & Patch Camelyon~\cite{Veeling2018qh}       & 2         &      \multirow{4}{*}{800/1,000}                 & \multirow{4}{*}{200}                & 32,768           \\
                          & EuroSAT~\cite{helber2019eurosat}              & 10        &                 &    & 5,400            \\
                          & Resisc45~\cite{cheng2017remote}             & 45        &                 &    & 6,300            \\
                          & Retinopathy~\cite{kaggle2015retinopathy}         & 5         &                 &    & 42,670           \\
 \midrule\multirow{8}{*}{Structured}                         & Clevr/count~\cite{johnson2017clevr}          & 8         &      \multirow{8}{*}{800/1,000}                 & \multirow{8}{*}{200}                  & 15,000       \\
                          & Clevr/distance~\cite{johnson2017clevr}       & 6         &                 &    & 15,000     \\
                          & DMLab~\cite{beattie2016deepmind}                & 6         &                 &    & 22,735           \\
                          & KITTI-Dist~\cite{geiger2013vision}           & 4         &                 &    & 711   \\
                          & dSprites/location~\cite{matthey2017dsprites}    & 16        &                 &    & 73,728           \\
                          & dSprites/orientation~\cite{matthey2017dsprites} & 16        &                 &    & 73,728           \\
                          & SmallNORB/azimuth~\cite{lecun2004learning}    & 18        &                 &    & 12,150     \\
                          & SmallNORB/elevation~\cite{lecun2004learning}  & 18         &                 &    & 12,150     \\ \midrule
                           \multicolumn{6}{c}{Few-shot learning}\\\midrule
 & Food-101~\cite{food}             & 101       & \multirow{5}{*}{1/2/4/8/16 per class} & 20,200 & 30,300            \\
                          & Stanford Cars~\cite{car}        & 196       &  & 1,635  & 8,041       \\
                          & Oxford-Flowers102~\cite{flower}    & 102       &  & 1,633  & 2,463             \\
                          & FGVC-Aircraft~\cite{fgvc}        & 100       &  & 3,333  & 3,333       \\
                          & Oxford-Pets~\cite{pets}          & 37        &  & 736   & 3,669    \\\midrule
                          \multicolumn{6}{c}{Supplementary experiments}\\\midrule & CIFAR100 (Full)~\cite{krizhevsky2009learning}             & 100       &     60,000             &  -  & 10,000    \\
                          &Pascal-Context~\cite{pascal}             & 59        &     4,996             & 5,104  & -   \\
\bottomrule[1.5pt]
\end{tabular}}\end{center}
\caption{\textbf{Statistics of used datasets.}
}
\label{tab:dataset}
\end{table*}

\begin{table*}[t]

\begin{center}
\scalebox{0.9}{
\begin{tabular}{cccccccc}
\toprule[1.5pt]
 &optimizer&batch size&learning rate&weight decay&\# epochs&lr decay&\# warm-up epochs\\ \midrule
 VTAB-1K&AdamW&64&1e-3&1e-4&100&cosine&10\\
 Few-shot learning&AdamW&64&5e-3&1e-4&100&cosine&10\\
 \bottomrule[1.5pt]
\end{tabular}}\end{center}
\caption{\textbf{Hyper-parameters.}
}
\label{tab:hyper}
\end{table*}
\end{document}


\title{Appendix}
\author{Shibo Jie, Haoqing Wang, Zhi-Hong Deng\\
School of Intelligence Science and Technology, Peking University\\
{\tt\small \{parsley,wanghaoqing,zhdeng\}@pku.edu.cn}
}

\ificcvfinal\thispagestyle{empty}\fi

\onecolumn
\section*{Appendix}
\setcounter{section}{0}
\section{Code}
Please refer to the anonymous GitHub link \href{https://anonymous.4open.science/r/Low-Bit-Adapters-9B41/README.md}{https://anonymous.4open.science/r/Low-Bit-Adapters-9B41/}
\section{Supplementary Experiments}
\subsection{Comparison with Other Quantization Methods}
We compare our quantization method with existing binary neural networks -- XNOR-Net~\cite{xnor} and IR-Net~\cite{irnet}. Similar to Bi-AdaptFormer, we use the quantization strategy of XNOR-Net and IR-Net to quantize (\ie, binarize) the weights of adapters to 1 bit, and keep the full-precision activation, called XNOR-AdaptFormer and IR-AdaptFormer, respectively. In experiments, we find that IR-AdaptFormer cannot be trained stably without Batch Normalization (BN)~\cite{bn}, so we add BN after each FC layer of the adapters. We set the hidden dimension $h=8$ for AdaptFormer, and $h=32$ for Bi-AdaptFormer, XNOR-AdaptFormer, and IR-AdaptFormer. 

As shown in Table~\ref{tab:irnet}, since IR-AdaptFormer uses additional BN and XNOR-AdaptFormer uses channel-wise scaling factors, their storage sizes are larger than that of Bi-AdaptFormer. IR-AdaptFormer results in significant performance degradation compared to AdaptFormer. We conjunct this is because IR-Net is designed for traditional convolutional networks equiped with BN and is not suitable for plug-in adapters in modern large-size vision architecture. XNOR-AdaptFormer also underperforms AdaptFormer, demonstrating the necessity of tailoring quantization strategy for adapters.

\begin{table}[h]
\centering
\setlength{\tabcolsep}{0.3pt}
\scalebox{0.88}{
\begin{tabular}{p{4.2cm}<{\centering}p{4.8cm}<{\centering}p{4.0cm}<{\centering}p{4cm}<{\centering}}
\toprule[1.5pt]

Backbone&Method&Avg. Acc.&Size (MB)\\\hline
\specialrule{0em}{1pt}{1pt}
\multirow{1}{*}{ViT-B/16}&Full&68.9&334.0\\
ViT-B/16&Linear&57.6&0\\
ViT-B/16&AdaptFormer&76.70&0.56\\
ViT-B/16&IR-AdaptFormer&72.13 \textcolor{red}{($\downarrow$ 4.57)}&0.14\\
ViT-B/16&XNOR-AdaptFormer&76.34 \textcolor{red}{($\downarrow$ 0.36)}&0.11\\
\rowcolor{lightgray}ViT-B/16&{Bi-AdaptFormer}&\textbf{76.97} \textcolor{ForestGreen}{($\uparrow$ 0.27)}&\bf0.071\\
\bottomrule[1.5pt]
\end{tabular}
}
\caption{\textbf{Average accuracy on VTAB-1K.}}
\label{tab:irnet}
\vspace{-8pt}
\end{table}

\subsection{Performance on Full CIFAR100}
In the VTAB-1K benchmark, each task only contains 1,000 training samples. We also conduct experiments on the full CIFAR100~\cite{krizhevsky2009learning}  dataset, which has a larger 60,000-image training set. Following~\cite{ssf}, we use a ViT-B/16 supervisedly pre-trained on ImageNet-21K with AugReg as backbone, and train the model for 100 epochs with batch size 128. We use $h=8$ for AdaptFormer and $h=32$ for Bi-AdaptFormer. All other settings are the same as in~\cite{ssf}.  As shown in Table~\ref{tab:cifar}, Bi-AdaptFormer still outperforms other baselines while using the smallest storage size. Compared to AdaptFormer, Bi-AdaptFormer brings 0.4\% performance improvement and 8$\times$ more storage efficiency.
\begin{table}[h]
\centering
\setlength{\tabcolsep}{0.3pt}
\scalebox{0.88}{
\begin{tabular}{p{4.2cm}<{\centering}p{4.8cm}<{\centering}p{4.0cm}<{\centering}p{4cm}<{\centering}}
\toprule[1.5pt]

Backbone&Method&Top-1 Acc.&Size (MB)\\\hline
\specialrule{0em}{1pt}{1pt}
\multirow{1}{*}{ViT-B/16}&Full\dag&93.82&334.0\\
ViT-B/16&Linear\dag&88.70&0\\
ViT-B/16&BitFit\dag&93.39&0.39\\
ViT-B/16&VPT-Shallow\dag&90.38&0.59\\
ViT-B/16&VPT-Deep\dag&93.17&1.76\\
ViT-B/16&SSF\ddag&93.92&0.78\\\cline{1-4}\specialrule{0em}{1pt}{1pt}
ViT-B/16&AdaptFormer&93.55&0.56\\
\rowcolor{lightgray}ViT-B/16&{Bi-AdaptFormer}&\textbf{93.95} \textcolor{ForestGreen}{($\uparrow$ 0.40)}&\bf0.071\\
\bottomrule[1.5pt]
\end{tabular}
}
\caption{\textbf{Accuracy on full CIFAR100.} \dag\ denotes results reported in~\cite{ssf}. \ddag\ denotes result re-implemented by directly re-runing the official code of~\cite{ssf} on our machines.}
\label{tab:cifar}
\vspace{-8pt}
\end{table}
\subsection{Semantic Segmentation}
As for the dense prediction, we apply our method on Segmenter~\cite{seg}. We use DeiT-B/16$_{384}$~\cite{deit} pre-trained on ImageNet-1K as encoder. Since each segmentation task tunes an individual decoder upon the pre-trained encoder, we use a single FC layer as a decoder which is much more lightweight than FCN~\cite{fcn} and MaskTransformer~\cite{seg}. We conduct experiments on Pascal-Context~\cite{pascal}. We evaluate three tuning paradigms: full fine-tuning, Adaptformer with $h=8$, and Bi-AdaptFormer with $h=32$. The models are trained for 50 epochs with batch size 64. As shown in Table~\ref{tab:seg}, Bi-AdaptFormer still outperform AdaptFormer in terms of both performance and efficiency.
\begin{table}[h]
\centering
\setlength{\tabcolsep}{0.3pt}
\scalebox{0.88}{
\begin{tabular}{p{4.8cm}<{\centering}p{4.8cm}<{\centering}p{4.0cm}<{\centering}p{4cm}<{\centering}}
\toprule[1.5pt]

Segmenter&Method&mIoU (SS)&Size (MB)\\\hline
\specialrule{0em}{1pt}{1pt}
Seg-B/16 (DeiT-B/16 + Linear)&Full&\bf52.61&335.1\\
Seg-B/16 (DeiT-B/16 + Linear)&AdaptFormer&51.57&0.56\\
\rowcolor{lightgray}Seg-B/16 (DeiT-B/16 + Linear)&{Bi-AdaptFormer}&{51.75} \textcolor{ForestGreen}{($\uparrow$ 0.18)}&\bf0.071\\
\bottomrule[1.5pt]
\end{tabular}
}
\caption{\textbf{Semantic segmentation on Pascal-Context.} ``Size'' denotes the size of trainable parameters in encoders. We report mIoU of single-scale inference on validation set. Each method also has a decoder of 0.17MB.}
\label{tab:seg}
\vspace{-8pt}
\end{table}

\section{Derivation of Eq. (8)}
If $i=k$,
\begin{align*}
    \frac{\partial \hat{w_i}}{\partial w_k}&=   \frac{\partial (\hat{w_i}'\cdot \sigma + \mu)}{\partial w_i}\\
    &=\sigma\cdot\frac{\partial \hat{w_i}'}{\partial w_i}+\hat{w_i}'\cdot\frac{\partial \sigma}{\partial w_i}+\frac{\partial \mu}{\partial w_i}\\    &=\sigma\cdot\frac{\partial {w_i}'}{\partial w_i}+\hat{w_i}'\cdot\frac{w_i'}{m}+\frac{1}{m}\\
    &=\sigma\cdot\frac{\partial\frac{w_i-\mu}{\sigma}}{\partial w_i}+\hat{w_i}'\cdot\frac{w_i'}{m}+\frac{1}{m}\\
        &=\sigma\cdot\frac{m-1-w_i'^2}{m\sigma}+\hat{w_i}'\cdot\frac{w_i'}{m}+\frac{1}{m}\\
        &=1+\frac{w_i'(\hat{w_i}' - w_i')}{m}\\
\end{align*}
\quad Otherwise,
\begin{align*}
    \frac{\partial \hat{w_i}}{\partial w_k}&=   \frac{\partial (\hat{w_i}'\cdot \sigma + \mu)}{\partial w_k}\\
    &=\sigma\cdot\frac{\partial \hat{w_i}'}{\partial w_k}+\hat{w_i}'\cdot\frac{\partial \sigma}{\partial w_k}+\frac{\partial \mu}{\partial w_k}\\    &=\sigma\cdot\frac{\partial {w_i}'}{\partial w_k}+\hat{w_i}'\cdot\frac{w_k'}{m}+\frac{1}{m}\\
    &=\sigma\cdot\frac{\partial\frac{w_i-\mu}{\sigma}}{\partial w_k}+\hat{w_i}'\cdot\frac{w_k'}{m}+\frac{1}{m}\\
        &=\sigma\cdot\frac{-1-w_i'w_k'}{m\sigma}+\hat{w_i}'\cdot\frac{w_k'}{m}+\frac{1}{m}\\
        &=\frac{w_k'(\hat{w_i}' - w_i')}{m}\\
\end{align*}

\section{Limitations and Broader Impact}
This paper is based on a series of empirical observations, including the flat local minima and approximately Gaussian weights of adapters. However, due to the difficulty to understand the complex learning behavior of deep neural networks, these observations are not supported theoretically in this paper.

Our paper reduces the size of trainable parameters required for fine-tuning ViT-B to several KBs without significant performance degradation. One may ask: \emph{is it necessary to make adapters so small?} Since the vision models are getting larger and larger, small businesses and individuals will not have enough computational resources to fine-tune the models, and the model owners will also be inclined to keep the pre-trained models private (\ie, language model GPT-3). Therefore, users will have to send labeled data to model owners (\ie, OpenAI) for paid fine-tuning (or adapter-tuning), and perform inference on the tuned models via online interface. Meanwhile, mobile devices can also benefit from centralized cloud computing. Then the problem arises. If the fine-tuned models are stored by a central server, it needs to store the fine-tuned models or smaller adapters for maybe millions of users. If the fine-tuned adapters (models are private) are stored by users, they need to send the adapters to the center server every time before inference on the models. Since the deeper and wider models also increase the size of adapters, adapter-tuning still leads to huge storage or transmission costs in view of the large base of users. Consequently, there are practical requirements to make adapters as small as possible.

\section{Experimental Details}
\subsection{Datasets}
See Table~\ref{tab:dataset}. 
\begin{table*}[h]

\centering
\scalebox{0.9}{
\begin{tabular}{clcccc}
\toprule[1.5pt]
                          & Dataset              & \# Classes & Train                    & Val   & Test            \\ \midrule
                          \multicolumn{6}{c}{VTAB-1K~\cite{vtab}}\\\midrule
\multirow{7}{*}{Natural} & CIFAR100~\cite{krizhevsky2009learning}             & 100       & \multirow{7}{*}{800/1,000}                 & \multirow{7}{*}{200}   & 10,000           \\
                          & Caltech101~\cite{fei2004learning}           & 102       &                 &    & 6,084            \\
                          & DTD~\cite{cimpoi14describing}                 & 47        &                 &    & 1,880            \\
                          & Oxford-Flowers102~\cite{flower}    & 102       &                 &    & 6,149            \\
                          & Oxford-Pets~\cite{pets}          & 37        &                 &    & 3,669     \\
                          & SVHN~\cite{netzer2011reading}                 & 10        &                 &    & 26,032            \\
                          & Sun397~\cite{xiao2010sun}               & 397       &                 &    & 21,750           \\
\midrule\multirow{4}{*}{Specialized}                          & Patch Camelyon~\cite{Veeling2018qh}       & 2         &      \multirow{4}{*}{800/1,000}                 & \multirow{4}{*}{200}                & 32,768           \\
                          & EuroSAT~\cite{helber2019eurosat}              & 10        &                 &    & 5,400            \\
                          & Resisc45~\cite{cheng2017remote}             & 45        &                 &    & 6,300            \\
                          & Retinopathy~\cite{kaggle2015retinopathy}         & 5         &                 &    & 42,670           \\
 \midrule\multirow{8}{*}{Structured}                         & Clevr/count~\cite{johnson2017clevr}          & 8         &      \multirow{8}{*}{800/1,000}                 & \multirow{8}{*}{200}                  & 15,000       \\
                          & Clevr/distance~\cite{johnson2017clevr}       & 6         &                 &    & 15,000     \\
                          & DMLab~\cite{beattie2016deepmind}                & 6         &                 &    & 22,735           \\
                          & KITTI-Dist~\cite{geiger2013vision}           & 4         &                 &    & 711   \\
                          & dSprites/location~\cite{matthey2017dsprites}    & 16        &                 &    & 73,728           \\
                          & dSprites/orientation~\cite{matthey2017dsprites} & 16        &                 &    & 73,728           \\
                          & SmallNORB/azimuth~\cite{lecun2004learning}    & 18        &                 &    & 12,150     \\
                          & SmallNORB/elevation~\cite{lecun2004learning}  & 18         &                 &    & 12,150     \\ \midrule
                           \multicolumn{6}{c}{Few-shot learning}\\\midrule
 & Food-101~\cite{food}             & 101       & \multirow{5}{*}{1/2/4/8/16 per class} & 20,200 & 30,300            \\
                          & Stanford Cars~\cite{car}        & 196       &  & 1,635  & 8,041       \\
                          & Oxford-Flowers102~\cite{flower}    & 102       &  & 1,633  & 2,463             \\
                          & FGVC-Aircraft~\cite{fgvc}        & 100       &  & 3,333  & 3,333       \\
                          & Oxford-Pets~\cite{pets}          & 37        &  & 736   & 3,669    \\\midrule
                          \multicolumn{6}{c}{Supplementary experiments}\\\midrule & CIFAR100 (Full)~\cite{krizhevsky2009learning}             & 100       &     60,000             &  -  & 10,000    \\   
                          &Pascal-Context~\cite{pascal}             & 59        &     4,996             & 5,104  & -   \\ 
\bottomrule[1.5pt]
\end{tabular}}
\caption{\textbf{Statistics of used datasets.}
}
\label{tab:dataset}
\end{table*}

\subsection{Pre-trained Backbones}
See Table~\ref{tab:pt}.

\begin{table*}[h]

\centering
\begin{minipage}{\linewidth}
\centering

\scalebox{1}{

\begin{tabular}{lccc}
\toprule[1.5pt]
Model&Pre-training Dataset&Size (M)&Pre-Trained Weights
\\\midrule
ViT-B/16~\cite{vit}&ImageNet-21K&85.8&\href{https://storage.googleapis.com/vit_models/imagenet21K/ViT-B_16.npz}{checkpoint}
\\
Swin-B~\cite{swin}&ImageNet-21K&86.7&\href{https://github.com/SwinTransformer/storage/releases/download/v1.0.0/swin_base_patch4_window7_224_22k.pth}{checkpoint}
\\
ConvNeXt-B~\cite{convnext}&ImageNet-21K&87.6&\href{https://dl.fbaipublicfiles.com/convnext/convnext_base_22k_224.pth}{checkpoint}\\
AugReg ViT-B/16~\cite{vit}&ImageNet-21K&85.8&\href{https://storage.googleapis.com/vit_models/augreg/B_16-i21k-300ep-lr_0.001-aug_medium1-wd_0.1-do_0.0-sd_0.0.npz}{checkpoint}
\\
DeiT-B/16$_{384}$~\cite{convnext}&ImageNet-1K&86.1&\href{https://dl.fbaipublicfiles.com/deit/deit_base_patch16_384-8de9b5d1.pth}{checkpoint}\\
 \bottomrule[1.5pt]
\end{tabular}
}
\end{minipage}

\caption{\textbf{Pre-trained backbones.}
}
\label{tab:pt}
\end{table*}

\subsection{Code Implementation}
We use \href{https://pytorch.org/}{\emph{PyTorch}} and \href{https://rwightman.github.io/pytorch-image-models/}{\emph{timm}} to implement all experiments on NVIDIA RTX 3090 GPUs.

\subsection{Data Augmentation}
\subsubsection{VTAB-1K} Following \cite{vpt}, we just resize the images to $224\times224$.

\subsubsection{Few-shot learning} Following \cite{noah}, for training samples, we use color-jitter and RandAugmentation; for validation/test samples, we resize them to $256\times256$, crop them to $224\times224$  at the center, and then normalize them with ImageNet's mean and standard deviation.
\subsubsection{Full CIFAR100} Following~\cite{ssf}, we use a strong augmentation in the fine-tuning setting of~\cite{vit}. Please refer to the official code of~\cite{vit}.

\subsubsection{Semantic Segmentation} We completely follow the setting used in~\cite{seg}, which does mean substraction, random
resizing, random left-right flipping, and randomly crops large images and pad small images to 480×480.

\subsection{Hyper-parameters}
$s$ is searched from \{0.01, 0.1, 1, 10, 100\}. See Table~\ref{tab:hyper} for other hyper-parameters. We basically follow the hyper-parameters used by~\cite{noah}.
\begin{table*}[h]

\centering
\scalebox{0.9}{
\begin{tabular}{cccccccc}
\toprule[1.5pt]
 &optimizer&batch size&learning rate&weight decay&\# epochs&lr decay&\# warm-up epochs\\ \midrule
 VTAB-1K&AdamW&64&1e-3&1e-4&100&cosine&10\\
 Few-shot learning&AdamW&64&5e-3&1e-4&100&cosine&10\\
 \bottomrule[1.5pt]
\end{tabular}}
\caption{\textbf{Hyper-parameters.}
}
\label{tab:hyper}
\end{table*}
{\small
\bibliographystyle{ieee_fullname}
\bibliography{egbib}
}